# Creating Domain-Specific Translation Memories for Machine Translation Fine-tuning: The TRENCARD Bilingual Cardiology Corpus

Gokhan Dogru, Universitat Autònoma de Barcelona

gokhan.dogru@uab.cat, ORCID: 0000-0001-7141-2350

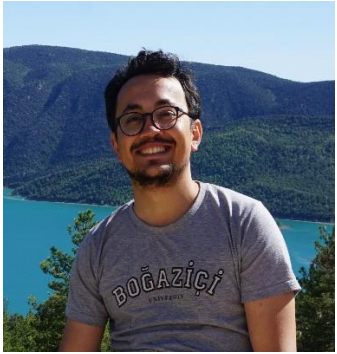

**Abstract**

This article investigates how translation memories (TM) can be created by translators or other language professionals in order to compile domain-specific parallel corpora , which can then be used in different scenarios, such as machine translation training and fine-tuning, TM leveraging, and/or large language model fine-tuning. The article introduces a semi-automatic TM preparation methodology leveraging primarily translation tools used by translators in favor of data quality and control by the translators. This semi-automatic methodology is then used to build a cardiology-based Turkish → English corpus from bilingual abstracts of Turkish cardiology journals. The resulting corpus called TRENCARD Corpus has approximately 800,000 source words and 50,000 sentences. Using this methodology, translators can build their custom TMs in a reasonable time and use them in their bilingual data requiring tasks.

## 1. Introduction

The rapid advancements in translation and localization technologies lead to a dilemma for today's professional translators in terms of technology use. On the one hand, there is the technological deflation paradigm: the most advanced proprietary translation platforms integrate vendor management, project management, computer-assisted translation (CAT) tools and machine translation (MT) features among others and provide a simple user interface that translators log into so as to work on translation segments without the need for controlling or mastering any individual technology involved. Web-based CAT tools such as Smartcat[1], Crowdin[2] and Lokalise[3] are some of the leading technology companies integrating all these features on platform and are managing the technological complexity on behalf of the translators at the backend. This consolidation of features is so widespread that it is argued in the European Language Industry Survey 2023 that "the integration of technologies into suites that combine most if not all required functionalities, it may soon be futile to look at individual tools" (ELIS 2023, p. 40). On the other hand, there is the technological inflation paradigm: the same wave of technological advancements and the rapid growth of the translation and localization industry have paved the way for the development of many new translation technologies including a new generation of CAT tools, terminology tools, MT systems and customization platforms etc. both in proprietary, and free and open-source versions. 2023 Nimdzi Technology Atlas (Nimdzi, 2023) report lists 920 language technology tools used in translation and localization industry. In line with the boom in the new technologies, Rothwell & Svoboda (2019) report in their survey that there was a notable increase in time devoted to teaching technology in translator training. While the technological deflation scenario seems to be more convenient for professional translators as it apparently minimizes the need to learn new technologies, it undermines the agency of translators (Moorkens, 2017), their ownership over translation data (Moorkens & Lewis, 2019b; Moorkens, 2022) and their control over the workflow, which, as a side effect, creates precarious working conditions for translators who work in large digital platforms on usually small chunks of translation projects without the guarantee of constant flow of job or income as highlighted by the concept of "uberization of translation" (Fırat, 2021). Although there is no deterministic reason for the technological deflation paradigm to aggravate the working conditions of the translators, in practice, it favors the owners of the platforms and accumulates the economic power on their side, making translators overdependent on a few platforms. The acknowledgment of the technological inflation paradigm, on the other hand, opens a plethora of technological possibilities for translators. Under this paradigm, while translators need to constantly learn new technologies, they can maximize their agency (Moorkens, 2017), be empowered as experts (O'Brien, 2012) and potentially get a bigger share of the efficiency gains by these technologies. And by having more control over the technologies and workflows, they may have more bargaining power over rates, among other things.

---

[1] Smartcat. https://www.smartcat.com/ [last access: 21.02.2024]
[2] Crowdin. https://crowdin.com/ [last access: 21.02.2024]
[3] https://lokalise.com/ [last access: 21.02.2024]

Within the above-mentioned framework, this study focuses on domain-specific translation memory (TM) creation for MT fine-tuning. The current tendency is for translators to work in CAT tools where MT systems are integrated (Farrell, 2022). They tend to use proprietary generic MT systems such as Google Translate and DeepL or custom MT systems provided by their clients, over which they have very limited control, if any. Since these MT systems are usually provided on the client side, and translators are expected to be the consumers of these systems, there tends to be an accompanying expectation of price discount, to the disadvantage of the translator (for a discussion of the effect of MT on pricing and productivity, see do Carmo, 2020). As an alternative to this scenario, we argue that translators are able to avoid this route by getting access to free and open-source tools, and by getting involved in different stages of MT system creation. This includes not only preparing parallel corpus in the form of TMs, but also evaluating the quality of training data, fine-tuning pre-trained NMT models, evaluating the quality of translation output and deploying their own MT systems in their CAT tools. For instance, free and open-source pre-trained NMT engines based on Marian NMT[4] system, and offered by the OPUS-MT project (Tiedemann & Thottingal, 2020), can be used in a desktop environment through the free and open-source OPUS-CAT[5] (Nieminen, 2021) software program that runs on Windows, thus allowing translators to connect MT engines to their CAT tools. This software also offers the possibility for local fine-tuning of these pre-trained NMT engines with the addition of custom translator data in the form of TMs. This makes the software even more useful, potentially, for translators as it allows them to customize MT engines with their own data. It may often be the case that TMs owned by translators are not big enough for fine-tuning the MT engines and, consequently, translators cannot benefit from the potential productivity gains that MT fine-tuning offers. Mikhailov (2022:224) observes that aligned corpora are still not sufficiently common, that these corpora are concentrated around a few language pairs and that they are available mostly as general domain data, not specific domain data. Hence, in the spirit of empowering translators in a technological inflation paradigm, we develop a semi-automatic TM creation procedure using a selection of tools familiar to translators.

---

[4] Marian NMT. https://marian-nmt.github.io/ (last access: 23.02.2024).
[5] Opus CAT. https://helsinki-nlp.github.io/OPUS-CAT/ (last access: 23.02.2024).

It is important to note that translators who typically work on CAT tools may not be familiar with programming languages (such as Python) or other command-line programs commonly used for web crawling and automatic bilingual corpus preparation. Hence, while the procedure for TM data compilation presented here is not as advanced as the methodologies using fully automatic web crawlers, it nevertheless allows translators to create TMs with a small amount of domain-specific, high quality data to be used in an MT fine-tuning procedure. In order to create very large monolingual, bilingual or multilingual corpora from web, the translator should ultimately resort to fully automatic web crawlers, as detailed in section 3, where different available tools are described. In this respect, it is important to note that the creator of OPUS-CAT, Nieminen (2021:291), highlights that fine-tuning with even a relatively small corpora of 10.000 sentences may provide quality improvements as opposed to a complete MT training from scratch that may require parallel corpora of more than millions of sentences (Pérez-Ortiz et al., 2022:148). In a study on productivity gains of NMT fine-tuning in the finance domain, Läubli et al. (2019) found that translators worked faster by 59.74% in the language pair German-French and by 9.26% in the pair German-Italian. Additionally, Gilbert (2020) fine-tuned Google AutoML with 1367 sentence pairs and observed that even with a very small fine-tuning data, the system already begins to learn from the style of the translator. In a comparative quality evaluation study on MT fine-tuning in the pairs English-Turkish, English-Spanish and English-Catalan in the localization domain, Dogru & Moorkens (2024) observed that human reviewers rated higher the fine-tuned engines across three language pairs in ranking, adequacy and fluency tasks. As Nieminen (2021:290) points out, domain fine-tuning strategies have been implemented at least since Koehn & Schroeder (2007) and numerous studies have shown different levels of quality gains.

## 2. Previous Work

TMs have been used in the translation industry since 1980s and the emergence of corpus based MT created an "an unanticipated connection" between TMs and this MT approach (Melby & Wrigh, 2015:675). This connection attributed a new role to TMs as "a type of parallel corpora" (Zanettin, 2012:169) that can be used for MT training and fine-tuning. The possibility of creating

TMs by aligning previously translated documents on a sentence level to provide parallel corpora for MT training and fine-tuning opened a new horizon for translators working in a computer-assisted translation (CAT) tool environment.

In the context of translation studies context, the use corpora in translation research and practice can be traced back to 1990s and early 2000s (Baker, 1993; Aston, 1999; Bowker & Pearson, 2002). These early works, as well as the subsequent ones by, for instance, Sánchez-Gijón (2009) and Marco & von Lawick (2009), focused on using corpora to support the development of translation competence and for research. Do-it-yourself (DIY) corpus procedures (Sánchez-Gijón, 2009) were also suggested in the abovementioned works. Yet corpus preparation for MT training and/or fine-tuning has not been commonly addressed in translation studies and has been performed mostly by computer scientists.

One of the earliest large-scale projects to create parallel corpora for MT was the Europarl Project, based on the proceedings of the European Parliament from the official website in 11 languages (Koehn, 2005). Later, Opus Corpus repository (Tiedeman, 2004) introduced and compiled more language pairs and toolkits that can be used for different purposes including MT training and fine-tuning. Moreover, several EU-funded large-scale projects contributed to increase the access to parallel corpora in different languages: Paracrawl (Esplà-Gomis et al., 2019) for all official EU languages; the EuroPat corpus (Heafield et al., 2022) for 6 official European languages paired with English: German, Spanish, French, Croatian, Norwegian, and Polish; MaCoCu (Bañón et al., 2022) for eleven low-resourced European languages; and Gourmet[6] for global under-resourced languages in the media domain. While these projects help solve the data quantity problem for many languages, there is still a need for high quality, domain-specific corpora in most languages, especially for the purpose of MT fine-tuning/domain adaption. For example, while it is possible to acquire the English-Spanish parallel corpus of the European Medical Agency (medical domain) or the United Nations (politics domain) through Opus Corpus, these corpora are not available for the English-Turkish language pair. Hence, there is still a need for corpus creation in different scales and domains.

---

[6] Global Under-Resourced Media Translation. https://gourmet-project.eu/ (last access: 23.02.2024)

One group of users who may be interested in TM preparation is the community of translators who can choose to deploy and adapt their own MT systems, whether through OPUS-CAT in their local Windows environment or through commercial platforms such as ModernMT[7] and Google AutoML Translation[8]. Translators can directly use the base MT engine of the relevant provider or fine-tune the engine by uploading their own TMs. If TMs are the product of their own translations, the size may be too small to effectively improve MT output to the extent that it would result in a significant productivity increase after fine-tuning (experimentation may be needed to decide upon the sufficient corpus size depending on the domain and language pair). Hence, another approach can be to create their own TM out of similar documents and/or websites.

One important concern in the literature has been the ownership of translation data (Moorkens, 2017 and 2022; Moorkens & Lewis, 2019a and 2019b. Translators already create TMs as a byproduct of their translation work. However, as Moorkens & Lewis (2019b) observe the ownership of this translation data is not clearly defined in many jurisdictions and, in practice, translators transfer their ownership to their clients when onsetting their collaborations to provide their services. These transfer usually means letting the client use this data not only for TM leveraging but also for all other uses including MT training and fine-tuning. This unsustainable practice exemplifies the technological deflation paradigm we outlined above. Moorkens & Lewis, (2019b) also question the sustainability of this approach and suggest treating translation data as a shared knowledge resource. On the other hand, when it comes to acquiring digital content from the Web for MT fine-tuning or for other purposes, it should be considered that this act may constitute an infringement of copyrighted material. Websites usually include copyright warnings or licenses detailing the permitted use cases of their content and they may have no-robot policies to stop crawling. An anonymization toolkit such as the one created by MAPA Project[9] may be needed to solve this problem. With the advent of data-driven artificial intelligence systems such as ChatGPT that depend on huge amount of internet data, concerns about the data use permissions continue to be raised and jurisdictions such as the EU are preparing laws to regulate these uses.

---

[7] ModernMT. https://www.modernmt.com/ (last access: 23.02.2024).
[8] Google AutoML Translation. https://cloud.google.com/translate/automl/docs (last access: 23.02.2024).
[9] MAPA project: https://mapa-project.eu/ (last access: 12.12.2023)

The following section outlines a semi-automatic procedure to compile a domain-specific bilingual corpus from the web that can later be used for fine-tuning an MT engine, among other use cases.

## 3. Methodology for TM Creation

Translators work with online and offline CAT tools with varying degrees of complexity, and permissions to implement TMs, glossaries, and MT integrations. When they offer translation services coupled with MT (e.g. post-editing) or are allowed to use the MT engine of their choice, they may improve the quality of the MT results by fine-tuning the engine through their TM resources.

While the underlying toolkit for corpus creation from web has changed since 2000s and new tools have been introduced such as Bitextor and Bicleaner (Esplà-Gomis et al., 2019), the overall automatic procedure remains very similar today: resource selection, web crawling, document alignment, sentence alignment, and file-type conversions of the final corpus (see for example Koehn, 2005). Based on the above-mentioned studies, we outline the end-to-end process of TM creation in Figure 1. This workflow can be completely manual, semi-automatic or fully automatic. In our suggested procedure, we have a semi-automatic, translator-friendly design. Semi-automatic approach allows for revising the automatic sentence alignments and ensures their overall quality, though the resulting corpus size tends to be smaller compared to fully automatic approaches.

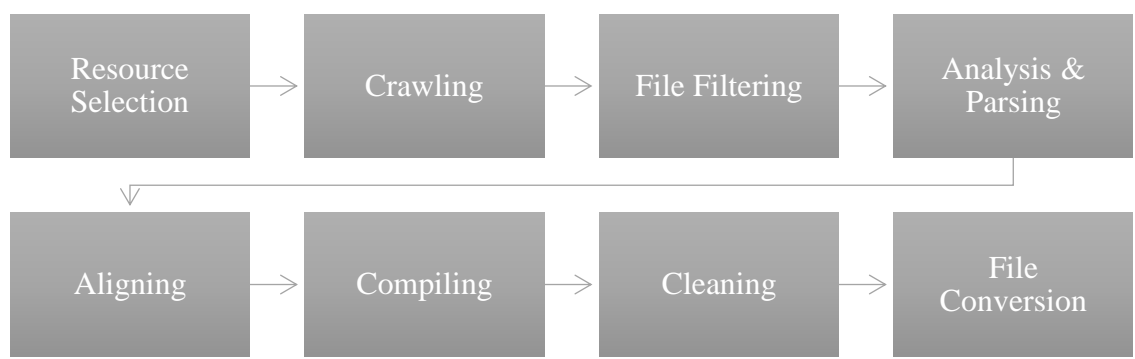

*Figure 1. Pipeline for TM creation.*

Below we will explain each step in the pipeline, the tools used and the reasons for using each tool In order to better illustrate the process we used this procedure to create a Turkish-English parallel corpus in the form of TM in the cardiology domain from bilingual abstracts from four Turkish cardiology journals. The practical example is described in detail in section 4.

### 3.1. Semi-Automatic Corpus Preparation Pipeline

The procedure below describes a process that begins with bilingual resource selection from the web, downloading web pages in different file formats with code pieces. A translator may skip some of these steps if they already have enough parallel documents in editable formats such as DOCX, XLXS, ODF etc. Furthermore, some tools may allow corpus cleaning before compiling or after file conversion, hence some steps may change order slightly. In a nutshell, as mentioned above, some steps such as resource selection, corpus cleaning etc. can be ignored or swapped depending on the available resources and toolkits as long as a final bilingual corpus is created.

### 3.1.1. Resource Selection

Resources for TM creation may vary from client-provided source and target files in different formats to bilingual physical publications and Web content. Clients may not always provide bilingual files in sufficient size, and digitalizing physical publications may be too time consuming. Although one should be careful about data ownership permissions and the data reliability, Web remains to be a popular data collecting alternative and is also the usual source for acquiring a parallel corpus. In his book on translation-focused corpora, Zanettin (2012:56) highlights, "The World Wide Web is the largest content distribution system and the most extensive and accessible repository of textual data". Using the Web, translators can search for bilingual websites with enough parallel content in their specific domain and decide to create parallel corpora in the form of TMs from it. The quality and quantity of this content plays a crucial role. The quality aspect does not only include translation quality but also the alignability of the content. Considering quantity, in general, the more specific the domain is, the less the possibility of acquiring large quantities of domain-specific data will be. Hence, it can be ideal to have available as large domain-specific parallel corpora as possible.

One approach for resource selection may be to consult the official websites from international organizations and government , and check whether they have content in the specific language pair. For example, the United Nations' Turkey website includes the UN Sustainable Development Goals in English[10] and Turkish[11]. This website can provide some parallel corpora for fine-tuning an engine for the legal/political domain. Professional associations such as engineering, architecture, medicine etc. usually have terminologically dense multilingual content. Zanettin (2012) provides a systematic overview of sources for translation driven corpora and Le Bruyn et al. (2022) approach the subject from a more technical perspective focusing on the representativeness of the corpus while collecting sources. Finally, the selection phase may be facilitated by the client who gives permission to the translator to freely use the content on their website. Or they may provide the bilingual files directly. The selection and curation of the files will greatly benefit from the expertise of the translator in the relevant domain.

Finally, when selecting resources from websites, data use permissions should be considered. Some websites have Creative Commons[12] symbols detailing the type of license the website provides and whether there are any limitations on the use of data.

### 3.1.2. Crawling

Crawling is the process of downloading content (textual, audiovisual and other types of content) from websites using a particular code or program. Crawlers or spiders can download a whole top-level domain, a list of URLs or content from a specific link. Users can set parameters to limit the time and amount of data to download and/or the type of files to be downloaded. HTTrack Website Copier[13] is a very common software tool used for this process. It is a multiplatform, free software with a graphical user interface (GUI), and one can download a website by simply pasting its link inside a search field in the software. It is possible to limit file sizes and types. In case there is no no-robot policy or limitation on the website, the download process begins to download all the content of the website. Another common tool that is easy-to-use is WGET[14]. It is a command-line tool that runs on the Linux operating system (OS). Most translators work on Windows but they can create a virtual machine using Oracle VM Virtual Box to run a free Linux OS[15] inside their Windows OS, and run WGET. It can be configured to download all or parts of websites, much like

---

[10] https://turkiye.un.org/en/sdgs/1 (last access: 12.12.2023)
[11] https://turkiye.un.org/tr/sdgs/1 (last access: 12.12.2023)
[12] https://creativecommons.org/share-your-work/cclicenses/ (last access: 12.12.2023)
[13] HTTrack Website Copier. https://www.httrack.com/ (last access: 23.02.2024)
[14] WGET. https://www.gnu.org/software/wget/ (last access: 23.02.2024)
[15] https://www.virtualbox.org/ (last access: 23.02.2024)

HTTrack. Usually one line of code is sufficient to trigger the download of a website. In some cases, WGET may prove to be more useful than HTTrack Website Copier depending on how the final downloaded files are displayed (see our experiment in the section below). Once the Linux Ubuntu Terminal is initiated, a code snipped such as the one below will be enough to start the download:

```
wget --mirror -p --convert-links --content-disposition --trust-server-names -P corpus http://khd.tkd.org.tr/
```

While HTTrack may change file names depending on the server settings on the website, WGET tends to keep the file names as they are, which is important when aligning web content. Once the download process is completed, there will be lots of unnecessary, non-textual or non-bilingual files downloaded. The translator needs to select the source and target documents to be aligned and remove all "noisy" files (if they have not done so already in the configuration of the crawling setting). At this step, the translator can enter their folder and manually remove any files such as images, configurations files or types of files with no content. If the operation of copying the files with textual content and pasting them into another folder requires less time than removing all other non-textual files, this operation can be opted. For example, if there are 100 HTML files with textual and 3000 files with images, configurations files or other non-textual content, copying these 100 HTML files and pasting them into another folder will be more convenient. Depending on the expected size of the fine-tuning corpora and file structure of the website, noise removal can be a quick step.

### 3.1.3. File Filtering

File filtering is a manual process of finding the relevant documents that can be aligned. In this manual filtering, the downloaded folder structure and file names play a crucial role, especially, when there are many files. Once there are equal number of source and target documents with consistent names, the analysis and parsing processes can begin.

### 3.1.4. Analysis and parsing

Parsing is the operation of separating textual strings from code in the context of translation. Documents downloaded from the Web come with their web format and not as plain texts. The usual file formats are HTML, XML, PHP etc. Hence, depending on the downloaded file type and the portion of the file to be aligned, translators can either use default file filters in their CAT tools such as OmegaT, memoQ, Trados Studio etc. or create custom filters using the filtering function of their CAT tools. To decide what type of filter/parser will be needed, translators need to open the files in a text editor such as NotePad++, which is a free option (see, for example, figure 2 in section 4.1.1). Once patterns are found in the analysis, a parsing strategy can be developed, which may include the use of some regular expressions (regex). The screenshot and regex text filter examples in Section 4.1.1. show how these operations are conducted. There may also be readymade external filters such as the Okapi Filters Plugin for OmegaT. This plugin can extend the capabilities of OmegaT to filter out advanced file types such as HTML and JSON. While in large language service companies, file processing is usually done by localization engineers, individual translators can also become familiar with filtering translatables from non-translatables, and CAT tools that provide a GUI for simple filter customization, such as memoQ, let translators perform this operation directly within a CAT environment.

### 3.1.5. Aligning

Alignment is a key operation for translators to leverage from previous bilingual files that are not in a TM format. Kraif (2002) provides the following definition for alignment:

> Aligning consists in finding correspondences, in bilingual parallel corpora, between textual segments that are translation equivalents. (p. 275)

Source segments can be paired with target segments automatically. For pairing the source and target documents sentence by sentence, the alignment software uses some parameters in the source and target sentences such as paragraph order, punctuation marks, inline tags, formatting, reference bilingual terminology, and relative word counts. One common command line tool used for this operation is HunAlign,[16] which automatically aligns sentences and provides a confidence score based on the parameters. Fully automatic corpus preparation pipelines use different versions of HunAlign to create aligned documents, since it is rather flexible. However, translators can use the alignment feature of their CAT tool. The free and open source OmegaT has an alignment feature

---

[16] Hunalign sentence aligner. https://github.com/danielvarga/hunalign (last access: 23.02.2024)

that can quickly align a pair of documents into a TMX[17] file, but it is not possible to align multiple files in a single step. MemoQ's LiveDocs feature allows alignment of multiple files and also has an interface for editing misalignments, which is a useful feature to improve alignment results. It is possible to adjust alignment parameters and a reference bilingual terminology to help the alignment process. LiveDocs also supports custom file filters before the start of the alignment. Trados Studio, which is also a common CAT tool among translators, includes a similar feature with similar capabilities. As a whole, what is crucial for an alignment component in a CAT tool is to be able to set parameters, to have an easy-to-use alignment editor, and to support multilingual file upload.

### 3.1.6. Compiling

In the context of this methodology presented here, compiling means combining all individually aligned file pairs into one single file including all of them. Once the corpora is aligned, the next step is to perform this compiling operation. This operation is optional, as sometimes it may be useful to keep the files separated if one wishes to allow for a fine-grained analysis of results in each of the files. If our choice is to compile all files into a single one, the use of the TMX format allows for interoperability between different CAT tools. Moreover, MT platforms such as ModernMT, KantanMT[18] and OPUS-CAT MT support TMX as an input file format for training and/or fine-tuning. Therefore, compiling aligned files into a single TMX file can be useful. MemoQ's LiveDocs feature allows to import the content of all aligned files into one single TMX file in one operation. As an extra step, once a TMX is created, it is possible to obtain two separate plain text files of the source side and target side easily through Okapi Rainbow.[19] Once a single file has been obtained, a cleaning process can be initiated in order to remove noise from the TMX file.

### 3.1.7. Cleaning

The quality of training data determines the resulting quality of the MT engine. For this reason, the training data (in the form of a TM) needs to be cleaned of any noisy data including inline tags,

---

[17] Translation Memory eXchange, a standard format for TM data exchange.
[18] KantanMT. https://www.kantanai.io/ (last access: 23.02.2024)
[19] Okapi Rainbow: https://okapiframework.org/wiki/index.php?title=Rainbow (last access: 23.02.2024)

non-textual characters, misaligned segments, duplicate sentences, incorrect encodings or very long sentences. Quality Assurance features of CAT tools can help detect such problems but there are also standalone tools such as Heartsome TMX Editor, Goldpan TMX/TBX Editor, and Okapi Framework tools such as CheckMate and Olifant that can be used to detect and remove these noisy contents. Depending on the quality of the raw TMX file, this manual step can take some time but it can improve the quality of NMT engines since NMT is quite sensitive to the quality of training data.

### 3.1.8. File Conversion

The last step may be necessary depending on the compiled file format. As mentioned above, the expected end product of TM for translators is a TMX file, but data used for training an MT system requires the use of a different format. Okapi Rainbow has a file conversion utility. TMX can be divided into 2 aligned plain text files which later can be uploaded for MT training or fine-tuning. Resulting plain text files can also be used for other operations such as monolingual term extraction or other text mining operations.

### 4. A practical example of TM preparation: the TRENCARD Corpus

Using the procedure detailed in section 3, we compiled the TRENCARD corpus[20], a Turkish-English TM for research and experimentation purposes. Below we explain step by step our procedure.

To begin with, we selected four cardiology journals which publish Turkish and English abstracts together with the scientific articles on their websites: *Archives of the Turkish Society of Cardiology, Turkish Journal of Cardiovascular Nursing, Turkiye Klinikleri Journal of Cardiology* and *Turkish Journal of Thoracic and Cardiovascular Surgery*. Abstracts in these academic cardiology journals were chosen based on achievable possible TM size, reliability of translation, existence of structured data for crawling as well as their terminological density. An important consideration while compiling a TM from the web was related to licenses. The journal abstracts from the *Archives of the Turkish Society of Cardiology, Turkish Journal of Cardiovascular Nursing* and *Turkiye Klinikleri Journal of Cardiology* are under the restrictive license of Attribution-NonCommercial 4.0 International (CC BY-NC 4.0)[21] while the abstracts from the *Turkish Journal of Thoracic and Cardiovascular Surgery* have a more permissive license of Attribution-NonCommercial 4.0 International[22] allowing to "redistribute, remix, transform, and build upon the material" without a commercial purpose. For this reason, although we explain our TM preparation methodology in practice on the abstracts from these four journals, we only redistribute the resulting TM from the fourth journal in GitHub. The size of this TM is more than 150,000 source words and approximately 14,000 sentences. For other details of this TM, refer to section 4.4. By replicating our methodology, as explained below, a comparable TM can be obtained from the abstracts of the remaining three journals subject to the license limitations.

---

[20] TRENCARD Corpus. https://github.com/gokhandogru/trencard

Abstracts in these Turkish medical journals are generally bilingual in Turkish and English. The remaining parts of the journal issues are mostly in Turkish. For this reason and due to license limitations, we chose to build the TRENCARD corpus from abstracts. However, abstracts typically contain 100-300 words and are short texts. Therefore, in order to build a TM of a relevant size we needed the compilation of a large number of abstracts. An advantage of using abstracts is that their content is typically very structured, which can help in parsing and alignment. A cardiology abstract starts with a section called "background" or "objective", then includes "methods", "results", "conclusion" and "keywords". Since word count is limited in abstracts, more terminology is used to express the content in a restricted number of words, which leads to terminological density. Considering that these four scientific journals are authoritative in their fields, it is convenient that they are published openly and that translations of the abstracts are of high quality (since they pass through a peer review process) and do not result from volunteer or crowd translations. In combination, these four cardiology journals reflect the last 30 years of cardiology study in Turkey, which makes the TM representative of Turkish cardiology domain. The oldest online issues are from 1990. The content of English and Turkish abstracts represents the common terminology used in this field.

**4.1. TM Preparation from *Archives of the Turkish Society of Cardiology***

---

[21] Attribution-NonCommercial-NoDerivs 2.0 Generic. https://creativecommons.org/licenses/by-nc-nd/2.0/ (last access: 23.02.2024)
[22] Attribution-NonCommercial 4.0 International https://creativecommons.org/licenses/by-nc/4.0/ (last access: 23.02.2024)

*Archives of the Turkish Society of Cardiology* "is a peer-reviewed journal that covers all aspects of cardiovascular medicine", is published in Turkish and English, and "accepts papers on a wide range of topics, including coronary artery disease, valve diseases, arrhythmias, heart failure, hypertension, congenital heart diseases, cardiovascular surgery, basic science and imaging techniques" (*Archives of the Turkish Society of Cardiology,* n.d.). The journal is published by the Turkish Society of Cardiology, which "is the scientific, nonprofit, nongovernmental organization of Turkish cardiologists, established on May 21st, 1963. Its 2360 members cover almost all the academicians and practitioners of cardiology and the related specialists in Turkey."[23] The online archive of the journal covers the period 1990-2024. As of 2005, each volume includes 8 issues and a varying number of supplements, which amount to more than 152 issues and 52 supplements until the current issue of 52 (1) from 2024. Issues have an editorial section, original articles, case reports, case images and a section called perspectives.

The website of the journal is in Turkish and English, and access to the contents of the articles is open. Furthermore, the website has a well-structured HTML that allows the processing of the pages easily. As explained below, we firstly used HTTrack Website Copier for downloading the content of the website. However, the file names of the downloaded HTML pages were not displayed in an ordered way; hence, we chose to use WGET instead. Below we explain our experience with these two website downloaders.

Using the main URL of the website (https://www.archivestsc.com/), the entire website was downloaded to our local computer by the default workflow of HTTrack Website Copier. All items inside the website were downloaded, including HTML files, style sheets, images, PDFs. Since folder structure was maintained, all English and Turkish abstracts were saved to separate folders. All unnecessary files were deleted, leaving only the HTML pages which include each abstract (there is one abstract in each HTML page). This resulted in 390 pages for each language. However, we observed that file names for each page were changed in a way that made it hard, if not impossible, to align files. The files names such as "jvi0a30", "jvi0aa5" did not appear to have a logical basis. Normally, on the website, each HTML page is hosted under a URL structure such as "website name + /jvi.aspx?un=TKDA-72699". Here the number after the "TKDA" attributes to the unique identity of each page. As we were not able to align the files properly in this setting, we searched for another website crawler and used WGET. WGET works on Ubuntu operating system. Since we worked in a Windows OS, we downloaded Oracle VirtualBox to be able to work on Ubuntu. After opening the terminal on Ubuntu, we typed this code to download all the website again:

```
wget --mirror -p --convert-links --content-disposition --trust-server-names -P TurkishCard https://www.archivestsc.com/
```

---

[23] Turkish Society of Cardiology. https://tkd.org.tr/en/menu/10/history (last access: 23.02.2024)

The name "TurkishCard" was given to the folder that would include all the files and subfolders. Similar to HTTrack Website Copier all content including javascript files, style sheets, images and other elements were downloaded and as before, folder structure was maintained. Yet, unlike the previous download, file names were kept as they were on the website. After the completion of the download, the folder was moved back to Windows and all unnecessary files were again removed, leaving only the files with abstracts. Each page including a Turkish abstract had a name such as "jvi.aspx_pdir=tkd&plng=tur&un=TKDA-00090" in which the section of "jvi.aspx_pdir=tkd&plng=tur&un=" was the same for all the Turkish abstracts. For the English abstracts, the names had a structure of "jvi.aspx_pdir=tkd&plng=eng&un=TKDA-24582" where "jvi.aspx_pdir=tkd&plng=eng&un=" was standard in all abstracts. This consistent pattern allowed us to use batch tasks. Windows 10 has an advanced command-line terminal called Power Shell[24] through which it is possible to batch rename files.[25] In order to simplify the handling of the files and match Turkish and English files, the Power Shell terminal for renaming was implemented by using the following command:

---

[24] Windows Power Shell. https://docs.microsoft.com/en-us/powershell/scripting/getting-started/getting-started-with-windows-powershell?view=powershell-7 (last access: 23.02.2024)

[25] In case the number of files is low and the translator wishes to avoid using the command line, manual matching or alignment of the files can be made. Yet, this process may take longer if there are many files to be aligned/matched.

```
PS D:\Academia\2019 - 2020 Thesis Completion Phase\PhD\Chapter 7_Methodology\Ready Corpora 2019\1. TKDA Journal\v2_TKDA Journal\www.archivestsc.com\uzun en> Dir | Rename-Item -NewName {$_.name -replace "jvi.aspx_pdir=tkd&plng=tur&un=","tr-"}
```

Through this command, we replaced "jvi.aspx_pdir=tkd&plng=tur&un=" with "tr-". Hence, the file names were abbreviated to short names such as "tr-TKDA-0900" which was easier to handle. The same process was repeated for the English abstracts. We observed that abstracts were repeated twice in both folders. Duplicates were removed. The total sum of abstracts was around 3920 in each language pair. There was a difference of 8 abstracts between the Turkish and English abstracts folders. Since finding the non-matching files was a time-consuming task , we changed the names of Turkish abstracts temporarily from "tr-" to "en-" so that they became equal to the English ones in terms of name, and a software called AllDup 4.2 was used to identify and delete the non-matching files. After finding and deleting the non-matching files, we renamed the Turkish abstracts again. Finally, we had 3918 files for each language; in other words, 3918 abstracts for Turkish and 3918 files for English.

After the file selection and alignment steps, sentence level alignment was initiated. This process started with analyzing and deciding on the most efficient strategy for parsing the files since we only needed the abstracts on each page. Manual copy-paste of each abstract into a plain text file is time consuming, therefore, a strategy for parsing each file and then aligning the files at a sentence level provides the possibility to save time. The free and open source CAT tool OmegaT has an alignment feature. However, it is not possible to batch-process files and parse a certain part of a file. HunAlign requires advanced technical skills for achieving our goals. For these reasons, we opted for memoQ's LiveDocs feature in which it was both possible to customize the file parsing filters and include only a certain part of the file, and thereafter, align these files directly. Once the automatic alignment was completed, an editor window was opened for editing the misaligned sentences in memoQ. Finally, when all sentence alignments are confirmed, all the sentences can be imported into a TM. To sum up, our steps hereinafter were as follows:

   1). Open Memoq's LiveDocs feature and create a new corpus

2). Add alignment pairs to the new corpus (All files in Turkish folder into one side and all files in English folder into another side)

3). Create a parsing filter for Turkish side and one for the English side

4). Select the correct language encoding

5). Start the filtering and aligning process

6). Check alignment editor for any mismatch

7). Import all sentences into a TM

8). Export this TM in TMX format

As mentioned in section 3.1, preparation steps may change order and since we use memoQ, alignment and parsing are swapped. When a HTML file is imported into memoQ, the program uses its default HTML filter.[26]

### 4.1.1. Parsing the Turkish abstracts

All text content (menu items, website related general texts etc.) will be imported with this default HTML filter. However, this can be changed, and either other default filters can be applied or custom filters can be created using regex text filter to extract only a specific part of a file. Moreover, more than one filter can be used to filter content from the code selectively. These consecutive filters are called cascading filters. Our cascading filters have a regex text filter (to extract only the relevant content) and a HTML text filter (to correctly visualize inline HTML tags in the file). In order to configure this filter for extracting the English and Turkish abstracts, we analyzed a few HTML files from our corpus. Note that for this filter to work on 3918 files, the structural pattern must be the same in all of them. Figure 2 shows the HTML structure of the file.

---

[26] We use the term "filter" with the same meaning as "parser".

Figure 2. HTML structure of one of the files including Turkish abstract as displayed in Notepad++ Editor. The grey area is the intended area to be extracted and the rest is not imported.

Our regex filter was able to filter the relevant text from the file. All abstracts had the same title structure between "<h2 class='journalArticleinTitletur'>" and "</h2>" tags. Secondly, content between "<br><p>" and "<hr noshade size=4 align=center color=#d3d3d3>" was the content to extracted. In other words, we wanted to filter the content of the abstract till the end of the keywords. Consequently, we had 2 regex rules for our first filter, which are as follows:

**Import only the following content:**

1- <h2 class='journalArticleinTitletur'>.*</h2> (Import everything between these two tags)

2- <br><p>.*<hr noshade size=4 align=center color=#d3d3d3> (Import everything between these two tags)

The ".*" regex symbols mean "every character". As can be observed from Figure 2, there are some inline tags that remain in between the extracted content which is perceived as plain text such as

"<br><b>METOD</b></br>". These characters both affect the sentence segmentation during alignment and may affect MT quality when used in training and/or fine-tuning. Hence, they had to be removed in the following steps. To facilitate this process, we applied a second filter (HTML filter) over the extracted content to recognize these characters as HTML tags. This allowed for removing HTML tags automatically in a tool with an automatic tag removal feature. The last important consideration in this step was the selection of the correct encoding for the language. Windows encoding (Windows-1254) provided the correct character set in this scenario[27]. And after the application of the cascading filter (regex text filter + HTML filter), we ended up having the segmented, clean Turkish abstract. Figure 3 shows how the filtered Turkish abstract and the English abstract look in LiveDocs Alignment Window.

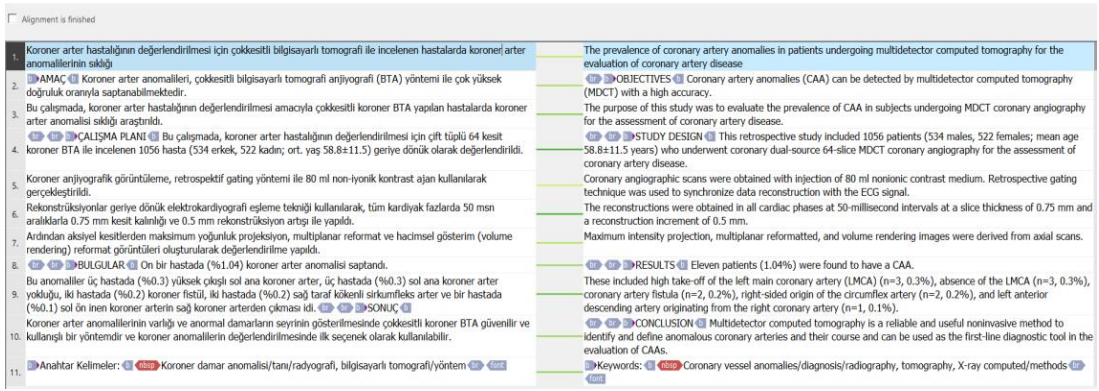

Figure 3. Alignment of source and target files in memoQ. Purple and red colored items are HTML tags.

### 4.1.2. Parsing the English abstracts

We followed a similar procedure for the English abstracts. Again, a cascading filter configuration is used for parsing. Only the first rule of the regex filter is slightly different:

**Import only the following content:**

1- <h2 class='journalArticleinTitleeng'>.*</h2> (Import everything between these two tags)

---

[27] This is discovered after a few trials with different characters sets such UTF-8.

2- <br><p>.*<hr noshade size=4 align=center color=#d3d3d3> (Import everything between these two tags)

Note that above only "<h2 class='journalArticleinTitleeng'>" tag is different in the filter. Another difference is with the encoding. We selected Western European (Windows) encoding for the English files. Once the parsing filters were set for both languages, we initiated the alignment step. Figure 3 shows how two files are aligned in memoQ. In this case, there are no misalignments. However, there may be some misalignments that need to be edited in some cases. We conducted some tests to make sure this parsing and alignment methodology can be used for all the files. These tests involved importing 10 file pairs into memoQ, applying the filters, conducting the alignment, and then checking whether there were misalignments. Once we made sure that the majority of the segments were correctly aligned, a batch processing for aligning all the 3918 Turkish files with 3918 English files is conducted using the corresponding filters. The process was completed with 27279 aligned sentences (segments).

All these sentences were compiled in a TM and are exported as a TMX file. In order to remove any tags, special characters, or any other kind of noise, we further processed this TM using cleaning tools. Once we combined all the content of the 3918 Turkish abstracts and 3918 English abstracts within a single TMX file, we could start the cleaning step. TM maintenance tools such as Heartsome TMX Editor and Goldpan TMX/TBX Editor, and Okapi Framework tools such as CheckMate and Rainbow are useful for implementing batch cleaning operations such as removing inline HTML tags, inconsistent numbers, and duplicate sentences. Essentially, this process is similar to translation quality assurance performed by professional translators. Figure 4 shows all the checks implemented by Goldpan TMX/TBX Editor.

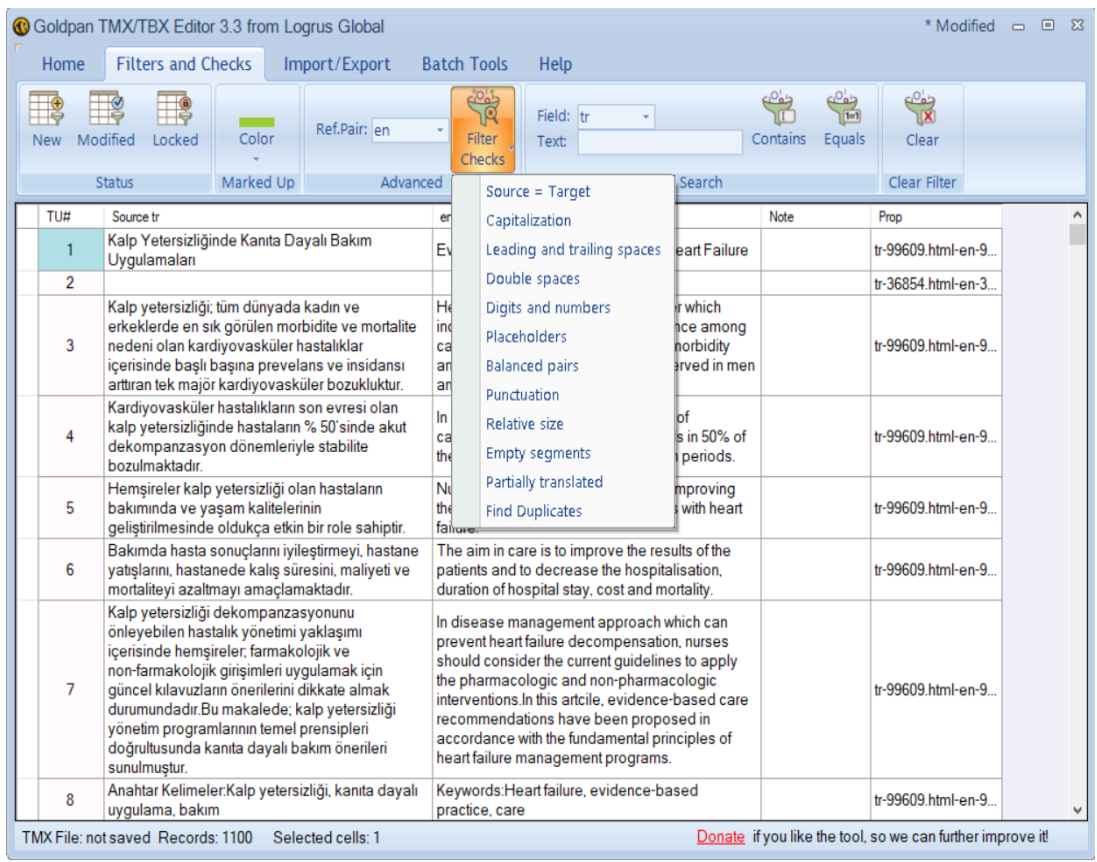

Figure 4. Goldpan TMX/TBX Editor has filter checks that allow 12 different checks to remove inconsistencies. Moreover, the Home tab includes a function for cleaning the tags in different formats. These options are used to optimize our TM.

We have only used GoldPan to clean our TM. When we completed cleaning, we ended up having a corpus in the form of a TMX file. Since MT platforms such as KantanMT, OPUS-CAT and MutNMT[28] support importing TMX files, we do not need to further process our file. In some MT tools, the files need to be prepared as two aligned plain text files. For example, older versions of MTradumàtica[29] require two distinct corpus files. Rainbow tool can be used for this operation. It can be noted that MT toolkits or systems such as KantanMT include internal operations for cleaning the TM automatically similar to the cleaning procedures mentioned above.

After importing the TMX file into Rainbow, going to Utilities > Conversion Utilities > File Format Conversion > Parallel Corpus Files and executing the command will create an aligned Turkish plain text file and an aligned English text file with the desired encoding. Although not needed for training/fine-tuning purposes, we conduct this step to be able to analyze the files when required. In the table below, we summarize the steps and the tools that are used for preparing our first journal TM.

---

[28] MutNMT. https://ntradumatica.uab.cat/ (last access: 23.02.2024)
[29] MTradumàtica. https://mtradumatica.uab.cat/ (last access: 23.02.2024)

|    | Phase                | Tools                          |
|----|----------------------|--------------------------------|
| 1. | Crawling & Downloading | WGET                         |
| 2. | Analysis             | NotePad++                      |
| 3. | Parsing              | Memoq (Filters Features)       |
| 4. | Aligning             | Memoq (LiveDocs Feature)       |
| 5. | Compiling            | Memoq (Export to TMX feature)  |
| 6. | Cleaning             | GoldPan TMX/TBX Editor         |
| 7. | File Conversion      | Okapi Rainbow                  |

Table 1. The tools and steps that we have followed for building the TM.

Following these steps, we created a TM of 27279 sentences, 496327 source words[30] with a word/sentence rate of 18.19[31]. Below we display the profile of our TM together with some meta information about it.

| Journal Name      | *Archives of the Turkish Society of Cardiology* |
|-------------------|--------------------------------------------------|
| Domain            | Cardiology                                       |
| UNESCO Code       | 3205.01                                          |
| Source Word Count | 496327                                           |
| Target Word Count | 570082                                           |

---

[30] Word count based on the Statistics feature of MemoQ.
[31] Based on dividing the number of words by the number of sentences.

| Sentence Count | 27279 |
| --- | --- |
| Source Word / Sentence Rate | 18.19 |
| Target Word / Sentence Rate | 20.89 |

Table 2. The TM profile of the first journal as calculated by Memoq's Statistics feature.

Lastly, challenges encountered while preparing this TM can be mentioned. Our first trials of crawling resulted in randomly named files in HTTrack, or in a way that did not allow to grasp the naming pattern. In order to solve this problem, we transitioned to WGET in Ubuntu. Secondly, while the encoding of the HTML pages is in UTF-8, using this encoding yields noisy characters. Following some trial and error, we found the correct encoding for the TM. Some translations had not been made sentence by sentence into English, which resulted in misalignments. In order to minimize this, we used terms as anchors in memoQ so that alignment reliability could increase. We derived these terms from the "keywords" section of the abstracts. Manuel check of the sentence level alignments was minimally used since overall alignment is considered to be adequate and possible misalignments were detected by Goldpan and then removed.

We used the procedure explained in Table 1 for all the 4 journals with some minor changes. Therefore, in the next sections, we will briefly explain the preparation procedure by repeating the same steps one by one.

### 4.2. TM from *Turkish Journal of Cardiovascular Nursing*

*Turkish Journal of Cardiovascular Nursing* is also a journal published by Turkish Society of Cardiology, more specifically by the Cardiovascular Nursing Technicians Working Group. It is "an Open Access, peer-reviewed e-journal that considers scientific research, case reports, reviews, translations, letters to the editor, news and abstracts presented at the National Congress of Cardiology"[32]. The topics covered include "the field of coronary artery disease, valvular heart disease, arrhythmias, heart efficacy, hypertension, congenital heart disease and all articles related to the coronary intensive care nursing."[33] The website of the journal, being from the same society,

---
[32] http://khd.tkd.org.tr/EN/about (last access: 23.02.2024)
[33] http://khd.tkd.org.tr/EN/about (last access: 23.02.2024)

has a similar design to the previous; hence, we utilize a similar procedure for this journal. We crawled the website with WGET, selected the relevant files, analyzed them, and determined the parsing filter that will use and implemented it. We use a cascading filter which has regex text filters and HTML filter just like we have done in the first journal. The regex for Turkish abstract includes only one rule:

**Import only the following content:**

1. <font color=#515151><h2>.*<br></font><br><hr noshade size=4 align=center color=#d3d3d3><font color=#515151><h2>

The reason why only one rule is used is that the structure of the HTML allowed for easier parsing. And the above tags are only included once in the file. If they were to be included more than once it would be impossible to use these rules. The example of the fourth journal presented in section 4.4. will better this. After the regex text filter, an HTML filter is applied too. The cascading filter for the English abstract includes a regex filter and one HTML filter. Regex text includes one 1 rule:

**Import only the following content:**

1. <font color=#515151><h2>.*<br></font><br><hr noshade size=4 align=center color=#d3d3d3><font color=#515151><h2>

As can be observed, the same parsing filter rule is used for both cases. The only difference between the English and Turkish files is that the order of the abstracts is changed. In the case of Turkish files, the Turkish abstract is above while in the case of English files, the English abstract is above. Hence, using the same filter configuration, we are able to filter and align the journal abstracts. After the alignment, we imported the content into a TM and exported it as a TMX file. In total, we have 1093 sentences, 17471 source words and 21019 target words.

| Journal Name | *Turkish Journal of Cardiovascular Nursing* |
|---|---|
| Domain | Cardiology |
| UNESCO Code | 3205.01 |
| Source Word Count | 17471 |
| Target Word Count | 21019 |
| Sentence Count | 1093 |
| Source Word / Sentence Rate | 15.98 |
| Target Word / Sentence Rate | 19.23 |

Table 3. TM profile of the second journal as calculated by memoQ's Statistics feature.

We also follow a cleaning procedure similar to the one used for the TM derived from the journal in the anterior section and remove the tags from the TMX files as well as unnecessary content through GoldPan. This TM preparation does not have any challenge. The only problem is that it is relatively small. However, its content is useful for the TRENCARD corpus.

### 4.3. TM Preparation from *Turkiye Klinikleri Journal of Cardiology*.

*Turkiye Klinikleri Journal of Cardiology* is another cardiology journal focusing on research publication in Turkey. It was published between 1988 and 2005. Its archive is distributed under an Attribution-NonCommercial-NoDerivatives 4.0 International (CC BY-NC-ND 4.0) Creative Commons license. After crawling its website by WGET, we develop a strategy to parse the files. We obtained 1018 files (abstracts). In this case, the files are downloaded with their original article title names without number reference. Hence, file alignment is not possible. However, each file includes both the Turkish and English abstracts. Using only one file in parsing step, we are able to divide the source and target content and, and then align them sentence by sentence. In a similar setup, a cascading filter with a regex text filter and an HTML filter is used. The regex for Turkish abstracts includes only one rule:

**Import only the following content:**

1. <div class="summaryMain"><b>ÖZET<br></b>.*</div>

And the regex for English abstracts also includes only one rule:

**Import only the following content:**

1. <div class="summarySub"><b>ABSTRACT<br></b>.*</div>

The abstract content is between tags that are occurring only once in the file. Otherwise, this kind of extraction would not be possible. The fourth journal will illustrate this case.

| Journal Name | *Turkiye Klinikleri Journal of Cardiology* |
|---|---|
| Domain | Cardiology |
| UNESCO Code | 3205.01 |
| Source Word Count | 118314 |
| Target Word Count | 133997 |
| Sentence Count | 7384 |
| Source Word / Sentence Rate | 16.02 |
| Target Word / Sentence Rate | 18.14 |

Table 4. TM profile of the third journal as calculated by Memoq's Statistics feature.

After parsing and alignment, we obtain our TMX file and clean it from HTML tags. The results yields 7384 sentences, 118314 source words, 133997 target words as it is displayed in the table above.

### 4.4. TM from *Turkish Journal of Thoracic and Cardiovascular Surgery*

The *Turkish Journal of Thoracic and Cardiovascular* Surgery is the last journal that we crawled for TM preparation. It is also a journal published in Turkish and English:

> *Turkish Journal of Thoracic and Cardiovascular Surgery* is an international open access journal which publishes original articles on topics in generality of Cardiac, Thoracic,

Arterial, Venous, Lymphatic Disorders and their managements. These encompass all relevant clinical, surgical and experimental studies, editorials, current and collective reviews, technical know-how papers, case reports, interesting images, How to Do It papers, correspondences, and commentaries.[34]

It is also licensed under Creative Commons Attribution-NonCommercial 4.0 International License. The website of the journal has a similar structure to the first two journals. However, in this case, the content cannot be extracted using the cascading filter (regex text filter and HTML filter) that is used before since the content of the abstracts is not included between unique tags or texts. For example, the tag "<div class="col-lg-12 col-md-12 col-sm-12 col-xs-12 makale-ozet">" is repeated several times; hence, when we try to parse the content between this tag and another tag, the tool cannot decide where to start the parsing. Different combinations of regex text filters are experimented with; however, it is not possible to extract content with this method. Since a typical abstract file in this journal does not include too much noisy content, we decide to use a default HTML filter. The noisy content that has to be imported includes menu items, the journal description in English as well as some meta data about how many times an abstract is viewed, authors etc. Since these contents are going to be both in the source segment and the target segment without change, they can be removed by using the "Remove Duplicate" function in GoldPan TMX editor. To sum up, we parse the abstracts with the HTML filter, align them and export them into TMX. Then in GoldPan, we remove the tags, the duplicated segments and, in the end, only the abstract sentences remains. This leads to spending a longer time in the cleaning phase. However, the result is the same. In the end, we have 13937 sentences, 155934 source words, 182284 target words.

| Journal Name | *Turkish Journal of Thoracic and Cardiovascular Surgery* |
|---|---|
| Domain | Cardiology |
| UNESCO Code | 3205.01 |
| Source Word Count | 155934 |

---

[34] http://tgkdc.dergisi.org/static.php?id=4

| | |
|---|---|
| Target Word Count | 182284 |
| Sentence Count | 13937 |
| Source Word / Sentence Rate | 11.18 |
| Target Word / Sentence Rate | 13.07 |

Table 5. TM profile of the fourth journal as calculated by Memoq's Statistics feature.

It can be observed that it is possible to achieve the same TM by concentrating on different steps of the TM preparation procedure, by first analyzing the files in detail, knowing the specificities each tool in the stack , and by implementing each steps accordingly.

**5 TRENCARD Corpus Compiled.**

Following the TM preparation procedure, we obtain 4 corpora on cardiology in TMX. Despite the fact that multiple files can be combined into a single TMX file, keeping them as separate files allows for a more fine-grained analysis of the results for each TM. Besides, as mentioned in section 4, due to license limitations, it is only possible to redistribute the resulting TM from the abstracts of the *Turkish Journal of Thoracic and Cardiovascular Surgery,* which amounts to approximately 20% of the whole study corpus.

Table 6 shows the compiled profile of TRENCARD. We created a TM of 49693 sentences, 788046 source words and 907382 target words. The source word/sentence count gives an idea about the average length of sentence in the TM. The TRENCARD Corpus with content from all four journals is available on Google Drive and access for research purposes is subject to permission by the author[35].

| Name | **TRENCARD CORPUS** |
|---|---|

---

[35] TRENCARD Corpus with all four corpora. https://drive.google.com/drive/folders/1E5UasfHEO9Qn668zgu4n8-UTe5cl3WCH?usp=sharing. As it is mentioned in section 4, the TM with the *content of the Turkish Journal of Thoracic and Cardiovascular Surgery* is on GitHub and it can be re-used for noncommercial purposes including research.

| Domain | Cardiology |
|---|---|
| UNESCO Code | 3205.01 |
| Source Word Count | 788046 |
| Target Word Count | 907382 |
| Sentence Count | 49693 |
| Source Word / Sentence Rate | 15.85 |
| Target Word / Sentence Rate | 18.25 |

Table 6. TRENCARD Corpus words, sentence count and word/sentence counts as calculated by Memoq's Statistics feature.

## 5. Limitations and Discussion

The semi-automatic procedure explained above requires using more than one stand-alone tool. While translation tools are usually easy to use and are part of translator training (Rothwell & Svoboda, 2019), switching between different tools can still be time-consuming to a certain degree. In addition, automatic alignment may not yield correct alignment pairs and posterior manual alignment may require some time. Since CAT tools are not specifically made for batch alignment, they may lack some flexibilities that tools like HunAlign may have. Furthermore, if there are many files to be aligned, the alignment process may take several days.

In the case of a very narrow domain, even after performing all the steps in the methodology, the resulting domain-specific TM may still not be large enough to influence the output quality of a fine-tuned NMT engine and further corpus compilation may be needed. Ramírez-Sánchez (2022:174) describes all the technical steps of preparing domain-specific data for custom MT, highlights the necessity to have a "generous" amount of domain-specific data but admits that it is hard to predict an exact amount. Dogru & Moorkens (2024) found that fine-tuning a pre-trained NMT engine with custom TMs in localization domain of 500,000 source words (69,500 sentences) already provided significant quality improvements (compared to the baseline engine) across three language pairs. Fine-tuning studies with smaller amount of domain-specific data can be performed to evaluate the effectiveness of smaller amount of data. Nieminen (2021) and Balashov (2021) hint that as little as 10,000 sentences can be enough to get significant results. Ramírez-Sánchez (2022) also notes that even tiny amounts of data are now being considered in adaptive or incremental MT scenarios, indicating a shift towards recognizing the value of any amount of data for learning and customization of MT systems. Besides, the preliminary studies with fine-tuning with large language models (LLMs pave the way for attaining better MT quality with smaller amount of custom TMs. In this respect, see, for instance, Moslem *et al.* (2023), where 20,000 sentences improved MT quality in medical domain, or the introduction of new technologies such as GPTs[36].

The data permissions and ethical data use constitute another limitation of our study. TM creation efforts from the web must observe the copyrights and use permissions of online resources although this limits even more the available domain-specific data. The meticulous observance of data permissions, ownership statuses, and ethical guidelines in the assembly of parallel corpora or TMs from web sources is, in any case, imperative.

## 6. Conclusion

Translators can create a sizable parallel corpus in the form of a TM using a semi-automatic procedure with already available tools. Using the procedure outlined in this paper, we created a TM of 788,046 source words in a very specific domain in only four days. Based on a standard daily translation output of 2500 words and 20 working days in a month, we can estimate that a translator can roughly translate 600,000 words a year. This makes the TRENCARD corpus larger than the annual output of a translator. The use of this methodology can help translators compile large enough TMs in their specialization field and lets them fine-tune their locally installed NMT engines in user-friendly OPUS-CAT software, empowering them in their professional work with the use of state-of-the-art technology without the need to depend on generic proprietary MT systems. And since they will not need to train an MT engine from scratch, they will not need advanced technical skills or intensive computational power. This procedure can also be helpful to create parallel corpora in the form of TMs for low resource languages.

---

[36] Introducing GPTs. https://openai.com/blog/introducing-gpts (last access: 23.02.2024)

Whether in TM form or in other parallel corpus forms, domain-specific data will continue to be important for translators, especially with data-intensive technologies such as LLMs. For this reason, it would be interesting to introduce in the future a new tool, following the inflation paradigm, combining all the corpus preparation steps described in our study within a single GUI, thus accelerating this preparation phase.

**Acknowledgements**

This work was partly supported by the DESPITE-MT project "Description of Posteditese in Machine Translation," grant number PID2019-108650RB-I00 [MICINN] and by the European Union-NextGenerationEU grant in the framework of Margarita Salas Postdoctoral Grant. Moreover, this work is derived from my PhD research; I would like to thank Adria Martin Mor and Anna Aguilar Amat for their contribution to this research. I am also grateful to Dr. Joss Moorkens, for his comments and feedback on this article.